# A GA-based Window Selection Methodology to Enhance Window-based Multi-wavelet transformation and thresholding aided CT image denoising technique

Prof. Syed Amjad Ali
Professor and Head of ECE Department
Lords Institute of Engineering and
Technology
Himayathsagar, Hyderabad – 8
syedamjadaliphd@gmail.com

Dr. Srinivasan Vathsal
Principal
Bhaskar Engineering College
Yenkapally, Moinabad
Ranga reddy Dist

Dr. K. Lal kishore
Rector, Jawahar Lal Nehru
Technological University
Kukatpally
Hyderabad.

*Abstract*— **Image denoising is getting more significance, especially in Computed Tomography (CT), which is an important and most common modality in medical imaging. This is mainly due to that the effectiveness of clinical diagnosis using CT image lies on the image quality. The denoising technique for CT images using window-based Multi-wavelet transformation and thresholding shows the effectiveness in denoising, however, a drawback exists in selecting the closer windows in the process of window-based multi-wavelet transformation and thresholding. Generally, the windows of the duplicate noisy image that are closer to each window of original noisy image are obtained by the checking them sequentially. This leads to the possibility of missing out very closer windows and so enhancement is required in the aforesaid process of the denoising technique. In this paper, we propose a GA-based window selection methodology to include the denoising technique. With the aid of the GA-based window selection methodology, the windows of the duplicate noisy image that are very closer to every window of the original noisy image are extracted in an effective manner. By incorporating the proposed GA-based window selection methodology, the denoising the CT image is performed effectively. Eventually, a comparison is made between the denoising technique with and without the proposed GA-based window selection methodology.**

## I. INTRODUCTION

Digital images are pivotally involved in the routine applications like satellite television, magnetic resonance imaging and computer tomography. In addition, they are involved in the areas of research and technology, namely, geographical information systems and astronomy. Retrieving original images from incomplete, indirect and noisy images is a serious issue that scientists experience in the aforesaid fields [1]. When the images are captured by the sensors and transmitted in the channel, the noises are added to the images [2]. With the existence of noise, the image gets a mottled, grainy, textured or snowy appearance [3]. Hence, in recent years, an overwhelming interest has been noticed in the case of recovering an original image from noisy image [4]. The recovery of an image is possible by image denoising. Image denoising can be described as the process of determining the original image from a contaminated image by noise degradation [5].

Generally, image denoising is the action of eliminating undesirable noise from a noised image, by bringing back the image to its un-degraded ideal [6]. The Image denoising techniques can be classified as transform domain methods or spatial domain methods. The transform domain methods transform an image from the spatial domain into another domain (probably, frequency domain or wavelet domain) and suppress noise in the transform domain, whereas, in spatial domain methods, the noise is suppressed in the spatial domain itself [7]. However, the image denoising with multi-wavelet techniques is very effectual due to the potential of capturing the signal energy in a small number of transformation energy values. In comparison with other multi-scale representations, the multi-wavelet transformation offers better spatial and spectral localization of image.

The image denoising finds its applications in fields of medical imaging and preprocessing for computer vision [8]. Medical imaging acquisition technologies and systems bring in noise and artifacts in the images and they should be attenuated by denoising algorithms. The denoising process should not damage anatomical details pertinent to a clinical viewpoint [9]. As a matter of reason, it is hard to put forward a robust method for noise removal which functions well for diverse modalities of medical images [10]. CT is regarded as a general and vital modality in Medical Imaging which is used for clinical diagnosis and computer-aided surgery [11]. In recent years, there have been numerous methods developed and described in literature for denoising [12].

In spite of the existence of several image denoising algorithms over the years, finding a proper solution for noise suppression in situations involving low signal-to-noise ratios remains a complex task [13]. In the earlier work, an efficient denoising technique for CT images employing window-based Multi-wavelet transformation and thresholding has been presented [32]. There the multi-wavelet has been favored since it betters single wavelets by its characteristics particularly, orthogonality, short support, symmetry, and high degree of vanishing moments. The technique has denoised the CT images degraded by AWGN and enhances the quality of the image. However, a drawback persists in choosing the closer windows in the process of window-based multi-wavelet transformation and thresholding. Normally, the windows of the duplicate noisy image that are closer to each window of original noisy image are acquired by the inspecting them sequentially. This results in the possibility of missing out very closer windows and so enhancement is needed in the aforesaid process of the denoising technique.

Here, we propose a GA-based window selection methodology to incorporate the denoising technique. With the aid of the GA-based window selection methodology, the windows of the duplicate noisy image that are very closer to every window of the original noisy image are extracted in an effectual way. By incorporating the proposed GA-based window selection methodology, the denoising is carried out more successfully. Eventually, a comparison is made between the denoising technique with and without the proposed GA-based window selection methodology. The rest of the paper is organized as follows. Section II briefly reviews the recent research works in the literature and Section III gives a short introduction about the GA. Section IV explains the window selection methodology of the denoising technique proposed in the previous work. Section V describes the proposed GA-based window selection methodology with required illustrations and mathematical formulations. Section VI discusses about the implementation results and Section VII concludes the paper.

## II. RELATED WORKS

Lanzolla et al. [14] have evaluated the effect of different noise reduction filters on computed tomography (CT) images. Especially, they have presented a denoising filter on the basis of a combination of Gaussian and Prewitt operators. Simulation results have proved that their presented technique has enhanced the image quality, and then permitted to use low radiation dose protocol in CT examinations. Their work was carried out in association with "G.Moscati" Hospital of Taranto (Italy), that offered all the images and technical materials employed in the proposed algorithm. Bing-gang Ye and Xiao-ming Wu [15] have addressed that the prior detection of small hepatocellular carcinoma (SHCC) has significant clinic value, and wavelet denoising arithmetic research of SHCC CT image, on the basis of image processing technology has aided to diagnose the SHCC focus. In accordance with the wavelet coefficient correlation, their work has reduced the figures and eliminated the feeble or irrelated coefficient of noise of SHCC CT image, and finally removed the noise.

The objective of Jin Li et al. [7] was to lessen the noise and artifacts in the industrial CT image by anisotropic diffusion. Anisotropic diffusion algorithms which could maintain significant edges sharp and spatially fixed at the same time as filtering noise and small edges eliminated the noise from an image by altering the image through a partial differential equation. In conventional anisotropic diffusions which lead to the loss of image details and cause false contours, 4-neighborhood directions are employed generally except diagonal directions of the image. To remove the drawbacks of the conventional anisotropic diffusion methods, an anisotropic diffusion method for industrial CT image based on the types of gradient directions was presented. In their work, one parameter K is calculated first by the histogram of the gradient. Then Sobel operator was made use of to calculate the directions of gradient. The directions of the gradient were classified. Experimental results have revealed that their presented algorithm could eliminate noise and artifacts from industrial CT volume data sets that were better than the Gaussian filter and other traditional algorithm.

Hossein Rabbani [16] have presented an image denoising algorithm based on the modeling of coefficients in each sub-band of steerable pyramid employing a Laplacian probability density function (PDF) with local variance. That PDF was able to model the heavy-tailed nature of steerable pyramid coefficients and the empirically observed correlation between the coefficient amplitudes. Within that framework, he has described a method for image denoising based on designing both maximum a posteriori (MAP) and minimum mean squared error (MMSE) estimators, which has relied on the zero-mean Laplacian random variables with high local correlation. Despite the simplicity of his spatially adaptive denoising method, both in its concern and implementation, his denoising results has achieved better performance than several published methods such as Bayes least squared Gaussian scale mixture (BLS-GSM) technique that was a state-of-the-art denoising technique.

H.Rabbani et al. [17] have proposed noise reduction algorithms that could be employed to improve image quality in several medical imaging modalities like magnetic resonance and multidetector CT. The acquired noisy 3-D data were first transformed by discrete complex wavelet transform. Employing a nonlinear function, they have modeled the data as sum of the clean data plus additive Gaussian or Rayleigh noise. They employed a mixture of bivariate Laplacian probability density functions for the clean data in the transformed domain. The MAP and minimum mean-squared error (MMSE) estimators enabled them to effectively reduce the noise. In addition, they have calculated the parameters of the model using local information. Experimental results on CT images revealed that among their derived shrinkage functions, generally, BiLapGausMAP has given images with higher peak SNR.

Skiadopoulos et al. [18] have carried out a comparative study between a multi-scale platelet denoising method and the well-established Butterworth filter, which was employed as a pre- and post-processing step on image reconstruction. The comparison was performed with and/or without attenuation correction. Quantitative evaluation was executed by using 1) a cardiac phantom comprising of two different size cold defects, employed in two experiments done to simulate conditions with and without photon attenuation from myocardial surrounding tissue and 2) a pilot-verified clinical dataset of 15 patients with ischemic defects. Furthermore, an observer preference study was executed for the clinical dataset, based on rankings from two nuclear medicine clinicians. Without photon attenuation conditions, denoising by platelet and Butterworth post-processing methods outplayed Butterworth pre-processing for large size defects. Conversely, for the small size defects and with photon attenuation conditions, all the methods have showed similar denoising performance. Guangming Zhang et al. [19] have proposed an extended model for CT medical image de-noising, which employed independent component analysis and dynamic fuzzy theory. Initially, a random matrix was created to separate the CT image for estimation. Then, dynamic fuzzy theory was applied to set up a series of adaptive membership functions to produce the weights degree of truth. At last, the weights degree was employed to optimize the value of matrix for image reconstruction. By putting to practice their model, the selection of matrix could be optimized scientifically and self-adaptively.

Jessie Q Xia et al. [20] have employed the partial diffusion equation (PDE) based denoising techniques particularly for breast CT at various steps along the reconstruction process and it was noticed that denoising functioned better when applied to the projection data rather than the reconstructed data. Simulation results from the contrast detail phantom have proved that the PDE technique outplayed Wiener denoising and also adaptive trimmed mean filter. The PDE technique has improved its performance features in relation to Wiener techniques when the photon fluence was lowered. With the PDE technique, the sensitivity for lesion detection employing the contrast detail phantom declined by less than 7% when the dose was reduced to 40% of the two-view mammography. For subjective evaluation, the PDE technique was employed to two human subject breast data sets obtained on a prototype breast CT system. The denoised images had great visual characteristics with a considerable lower noise levels and enhanced tissue textures while retaining sharpness of the original reconstructed volume.

A. Borsdorf et al. [21] have proposed a wavelet based structure-preserving method for noise reduction in CT images that could be used together with various reconstruction methods. Their approach was on the basis of presumption that the data could be decomposed into information and temporally uncorrelated noise. The analysis of correlations between the wavelet representations of the input images enabled separating information from noise down to a certain signal-to-noise level. Wavelet coefficients with small correlation were reduced, while those with high correlations were supposed to symbolize structures and are preserved. The ultimate noise-suppressed image was reconstructed from the averaged and weighted wavelet coefficients of the input images. The quantitative and qualitative evaluation on phantom and real clinical data proved that high noise reduction rates of around 40% could be accomplished without considerable loss of image resolution.

### III. GENETIC ALGORITHM (GA)

The GA-based approaches have received considerable interest from the academic and industrial communities for coping with optimization problems that have proved to be difficult by employing conventional problem solving techniques [22][23] [24][25][26]. GAs are computing algorithms designed in correlation to the process of evolution [27], which was proposed in the 1970s in the United States by John Holland [28]. In GA, the search space comprises of solutions which are represented by a string identified as a chromosome. Each chromosome is composed of an objective function called fitness. In GA, the search space consists of solutions which are denoted by a string known as a chromosome. A collection of chromosomes together with their associated fitness is termed as the population. The population, at a particular iteration of the GA, is known as a generation [28] [29] [30].

GA begins to function with numerous possible solutions that are obtained from the randomly generated initial population. Then, it tries to find optimum solutions by employing genetic operators namely selection, crossover and mutation [30]. Selection is a process of selecting a pair of organisms to reproduce. Crossover is a process of swapping the genes between the two individuals that are reproducing. Mutation is the process of randomly modifying the chromosomes [27]. The main aim of mutation is re-establishing lost and exploring variety of data. In accordance with changing some bit values of chromosomes provide different breeds. Chromosome may be better or poorer than old chromosome. If they are poorer than old chromosome, then they are removed from selection step [31]. The process continues until a termination criterion is satisfied and so the GA can converge to an optimal solution.

### IV. WINDOW SELECTION METHODOLOGY IN THE DENOISING TECHNIQUE USING WINDOW-BASED MULTI-WAVELET TRANSFORMATION AND THRESHOLDING

Prior to detail the proposed GA-based window selection methodology for the CT image denoising technique [32], here, a brief description about the prevailing window-selection methodology used in the technique is given. Let, $I(x, y)$ be the original CT image and $I_{AWGN}(x, y)$ be the image affected by AWGN, where, $0 \le x \le M - 1$, $0 \le y \le N - 1$. The $I_{AWGN}$ is put to the first stage of the proposed technique, window-based thresholding. The window selection methodology to be described here is one of the components of the first stage of processing of the CT image denoisng technique. In the methodology, a replica of the $I_{AWGN}$, labeled as $I'_{AWGN}$, is generated. From $I_{AWGN}$ and $I'_{AWGN}$, a window of pixels are considered and put to multi-wavelet transformation. The process of extracting the windows from the image $I_{AWGN}$ is given in the Figure 1.

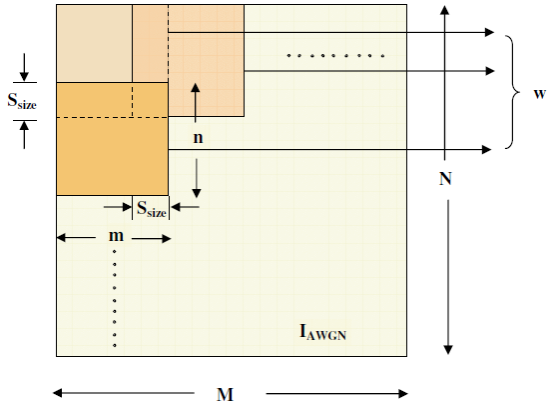

Figure 1. Process of extracting the windows from the given image IAWGN

In the Figure 1, $w$ indicates the window of pixels extracted from the image $I_{AWGN}$ and $S_{size}$ is the step size of the window. This is carried out throughout the image and so $w_i$ windows are attained, where, $0 \le i \le n_w - 1$. By the same way, it is also executed in the image $I'_{AWGN}$ and receives $w'_j$, $0 \le j \le n_w - 1$, where, $n_w$ represents number of windows. Then, the received window of pixels are transformed to multi-wavelet transformation domain as follows

$$W_i(a,b) = F_{GHM}(a,b) . w_i(a,b) . F^T_{GHM}(a,b) \qquad (1)$$

$$W'_j(a,b) = F_{GHM}(a,b) . w'_j(a,b) . F^T_{GHM}(a,b) \qquad (2)$$

where, $0 \le a \le m-1$, $0 \le b \le n-1$ and $m \times n$ indicates the window size. In (1) and (2) $F_{GHM}$ is the concatenated filter coefficient of GHM multi-wavelet transformation, $W_i$ and $W'_j$ are $w_i$ and $w'_j$ in multi-wavelet domain, respectively. For every $W_i$, $W'_j$ that are closer to $W_i$ are chosen based on L2 norm distance ($L2_{ij}$), which can be computed using (3),

$$L2_{ij} = \sqrt{\sum_{a=0}^{m-1}\sum_{b=0}^{n-1}(|W_i(a,b) - W'_j(a,b)|)^2} \qquad (3)$$

Using the $L2_{ij}$, the $W'_j$ windows that are closer to the $W_i$, $W'_{L2_{ij}}$ can be identified as $W'_{L2_{ij}} = W_{L2_{ij}} - \phi$, where, $W_{L2_{ij}}$ is given as

$$W_{L2_{ij}} = \begin{cases} W'_j & ; \quad if \quad L2_{ij} \le L2_T \\ \phi & ; \quad else \end{cases} \qquad (4)$$

Every $i^{th}$ window sets in $W'_{L2_{ij}}$ are sorted in ascending order based on their corresponding $L2_{ij}$. From the sorted window set, $n_c$ number of windows are chosen (for every $W_i$) and the remaining are omitted out, which leads to receive $W'_{L2_{ik}}$, where $0 \le k \le n_c - 1$. In the aforesaid window selection methodology, the time consumption is more. If the methodology is planned to be executed in less time, then it will lead to miss out closer windows. In order to overcome the drawback, we propose a GA-based window selection methodology, which selects much closer windows in a very less time.

## V. PROPOSED GA-BASED WINDOW SELECTION METHODOLOGY

Here, we propose a GA-based window selection methodology to replace the prevailing methodology performed in the process of window-based multi-wavelet transformation and thresholding in the denoising technique [32]. The proposed window selection methodology for the denoising technique using multi-wavelet transformation and window-based thresholding is depicted in the Figure 2. The methodology is made more effective by performing the mutation operation of the GA, adaptively.

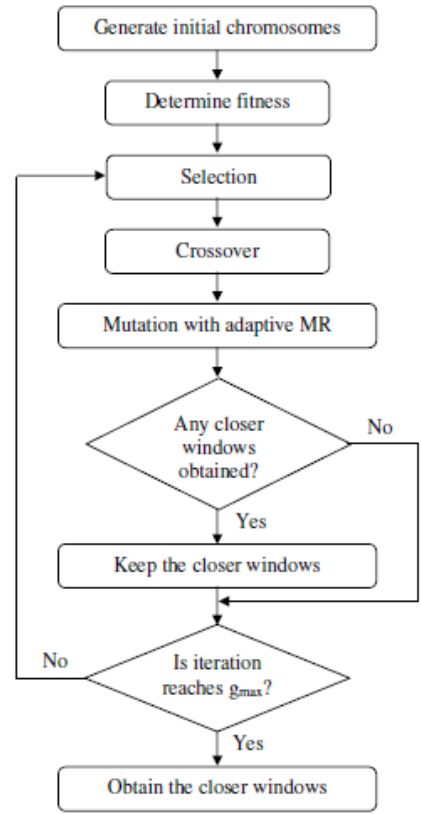

Figure 2. Process performed in the proposed GA-based window selection methodology

It is well known that the proposed GA-based window selection methodology is utilized to obtain $n_c$ number of windows, $W'_{L2_{ik}}$ ; $0 \le k \le n_c - 1$ that are closer to every $W_i$. Once the closer windows are identified, the further process of the denoising is continued using the obtained windows. The proposed methodology is comprised of five functional steps, namely, 1) generation of initial chromosomes, 2) Determination of fitness function, 3) Crossover and Mutation, 4) Selection of closer windows and 5) Termination criteria. They are described below in detail.

### A. Generation of initial chromosomes

In the methodology, as the first process, $n_g$ initial chromosomes, each of length $n_c$ are generated. The set representation of initial chromosomes are given as

$$\{R\}_{il} = \left\{ r_0, r_1, r_2, \cdots r_{n_c - 1} \right\}_{il} ; \quad 0 \le l \le n_g - 1 \qquad (5)$$

where, $\{R\}_{il}$ is the $l^{th}$ chromosome generated to obtain windows that are closer to the $i^{th}$ window of the original noisy image. Each gene of the generated chromosome $r_{ilk} \in \{R\}_{il}$ ; $0 \le k \le n_c - 1$, is an arbitrary integer generated within the interval $[0, n_w - 1]$ provided that the all the genes of each chromosome has to satisfy the condition $r_0 \ne r_1 \ne \cdots \ne r_{n_c - 1}$.

### B. Determination of fitness function

A fitness function decides whether the generated chromosomes are fit to survive or not, that can be given as

$$f_i(l) = \frac{1}{n_c} \sum_{k=0}^{n_c - 1} L2_{ilk} \qquad (6)$$

where, $f_i(l)$ is the fitness of the $l^{th}$ chromosome generated for the $i^{th}$ window and $L2_{ilk}$ is the $L2$ norm distance determined between the $w_i$ and the window indexed by the $k^{th}$ gene of the $l^{th}$ chromosome. The $L2_{ilk}$ is determined as follows

$$L2_{ilk} = \sqrt{\sum_{a=0}^{m-1} \sum_{b=0}^{n-1} \left( \left| W_i(a,b) - W'_{r_{ilk}}(a,b) \right| \right)^2} \qquad (7)$$

where, $W'_{r_{ilk}}$ is the window indexed by $r_{ilk}$ that is converted to multi-wavelet transformation domain as done in (1) and (2). From the $n_p$ generated chromosomes, $n_p / 2$ chromosomes that have minimum fitness are selected as best chromosomes and they are subjected to the genetic operations, crossover and mutation.

### C. Crossover and Mutation

Crossover and Mutation are the two major genetic operations which help the solution to converge soon. In the proposed methodology, double point crossover is selected to perform the crossover operation. In the double point crossover, two crossover points, $C_{p_1}$ and $C_{p_2}$ are chosen to meet a crossover rate of $CR$. In the crossover operation, the genes that are beyond the crossover points, $C_{p_1}$ and $C_{p_2}$, are exchanged between two parent chromosomes. Hence, $n_p / 2$ children chromosomes are obtained for the $n_p / 2$ parent chromosomes (that are selected as best chromosomes based on fitness function) from the crossover operation. Thus the obtained $n_p / 2$ children chromosomes are then subjected to the next genetic operation, Mutation.

The mutation operation to be performed, here, is effective as the mutation rate $MR$ is made adaptive with respect to the fitness function. The adaptiveness is accomplished by selecting the mutation points as well as the number of mutation points $n_{m_p}$ based on the fitness of the chromosomes. Hence, for each child chromosome, the mutation points and $n_{m_p}$ varies and they can be obtained as

$$m_{p_{il}}(k) = \begin{cases} m'_{p_{il}}(k) & ; if \; \sum_{k=0}^{n_c - 1} m''_{p_{il}}(k) = 0 \\ m''_{p_{il}}(k) = 0 \; ; otherwise \end{cases} \qquad (8)$$

where,

$$m'_{p_{il}}(k) = \begin{cases} 1 \; ; if \; \mathrm{M}_{L2_{il}} = k \\ 0 \; ; otherwise \end{cases} \qquad (9)$$

$$\mathrm{M}_{L2_{il}} = \underset{k \in [0, n_c - 1]}{\arg \max} \; L2_{il} \qquad (10)$$

$$m''_{p_{il}}(k) = \begin{cases} 1; \; if \; L2_{ilk} \ge L2_T \\ 0; \; otherwise \end{cases} \qquad (11)$$

The $n_{m_p}^{(i)}(l)$ is nothing but the number of unit values present in $m_{p_{il}}$, i.e., the mutation is performed at the $k^{th}$ gene only if $m_{p_{il}}(k) = 1$. The mutation is performed by changing the gene value by another arbitrary integer chosen at the interval $[0, n_w - 1]$. Hence, new $n_p / 2$ children chromosomes are obtained but that would satisfy the condition, $r_0^{new} \ne r_1^{new} \ne \cdots \ne r_{n_c - 1}^{new}$. If the condition is not satisfied, the mutation is performed in the corresponding child chromosome until it gets satisfied. Once, the mutation operation gets completed, the population pool is filled up by the selected best $n_p/2$ initial chromosomes and $n_p/2$ new children chromosomes. Hence, the population pool is comprised of $n_p$ chromosomes and they are subjected to the selection of closer windows.

### D. Selection of closer windows

The closer windows are selected by identifying the windows which has minimum L2-norm distance with the $i^{th}$ window as follows $\{R^{sel}\}_i << r_{ilk}$ ; $if\ L2_{ilk} < L2_T$. From the set $\{R^{sel}\}_i$, the closer $n_c$ windows are selected either by sorting the $\{R^{sel}\}_i$ elements in ascending order based on the corresponding $L2$ norm distance (if $|R^{sel}|_i > n_c$) or by selecting the $\{R^{sel}\}_i$ elements as the best closer windows (if $|R^{sel}|_i \le n_c$). Thus the selected elements occupy the set $\{R^{best}\}_i$. Now, with the $n_p$ chromosomes in the population pool, the process is repeated from Step 2 until it satisfies the termination criteria. At every iteration, the elements in the $\{R^{best}\}_i$ are updated, if any windows are obtained closer than the windows indicated by the $\{R^{best}\}_i$ elements. Hence, when every iteration gets completed, the closer $n_c$ windows are obtained rather than the windows obtained at the previous iteration.

### E. Termination Criteria

The process is repeated until the iteration reaches the maximum generation $g_{max}$. Once the iteration gets reached the $g_{max}$, then the $\{R^{best}\}_i$ is checked for the condition $|R^{best}|_i = n_c$. If this condition gets satisfied, the process is terminated and the $\{R^{best}\}_i$ are considered as the closer $n_c$ windows for the $i^{th}$ window, otherwise, iteration is continued for another $g_{max}$.

Thus, obtained $\{R^{best}\}_i$ is the $W'_{L2_{ik}}$ and it is subjected to the further steps of the denoising technique, thresholding, reconstruction and enhancement of the image [32].

## VI. Result and Discussion

The proposed window selection methodology has been implemented in the working platform of MATLAB (version 7.8). As described in denoising technique [32], $n_c = 16$ number of windows has to be selected for every $w_i$. Hence, in the proposed methodology, the gene length of $n_c = 16$ has been selected. The methodology has been initialized with a population size of $n_g = 10$ with a maximum generation of $g_{max} = 100$ and the each gene of the chromosome has been generated in the interval [0,3965] (i.e. $n_w = 63 \times 63$). In the genetic operations, crossover has been performed by selecting the crossover points as $C_{p1} = 5$ and $C_{p2} = 12$ and so $CR = 0.5$ has been met by the operation. As the mutation has been made adaptive, the $n_{m_p}$ and so $MR$ change dynamically. Once the process has been terminated, closer $n_c$ windows have been obtained for every $w_i$. This has been subjected to the further process of CT image denoising technique using window-based multi-wavelet transformation and thresholding. The proposed methodology has been evaluated by giving some CT images that are affected by AWGN at different levels ($\sigma = 10, 20, 30, 40\ and\ 50$). The results obtained for the noisy image, denoised image by the denoising technique with and without the proposed GA-based window selection methodology is given below.

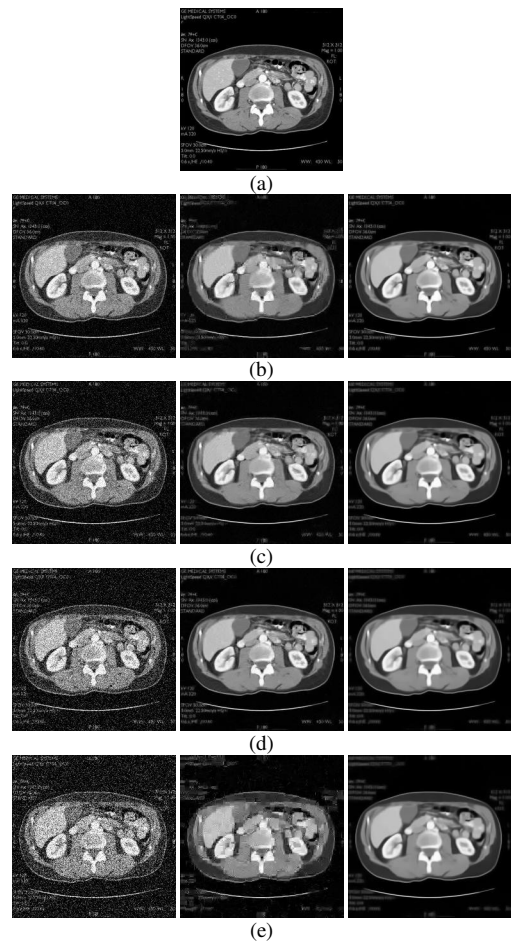

Figure 3. (a) original CT image of abdomen, (b), (c), (d) and (e) the image affected by AWGN at the levels of σ=20,30,40 and 50, respectively and the corresponding denoised image using the denoising technique with and without proposed GA-based window selection methodology.

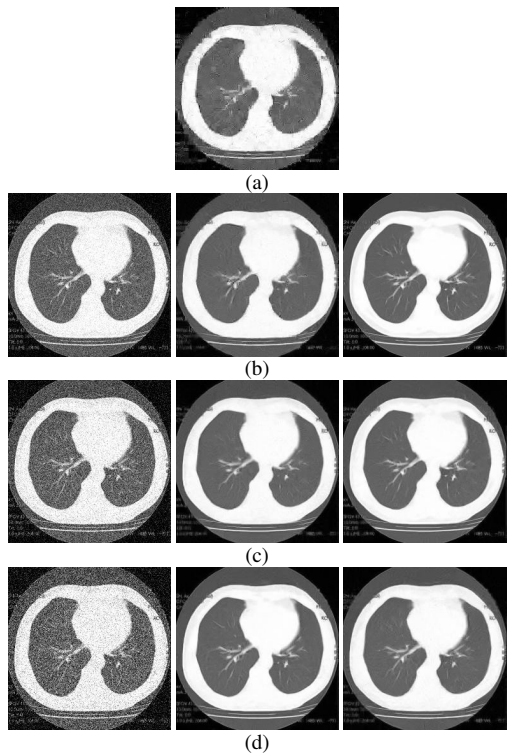

(a)

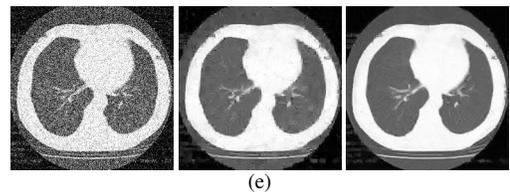

(e)

Figure 4. (a) original CT image of Thorax, (b), (c), (d) and (e) the image affected by AWGN at the levels of σ=20,30, 40 and 50, respectively and the corresponding denoised image using the denoising technique with and without proposed GA-based window selection methodology.

It can be visualized in Figure 3 and 4 that the denoising technique with the proposed GA-based window selection methodology has outperformed the technique without the methodology. A comparison is provided between the denoising technique with and without the proposed window selection methodology by comparing the PSNR values of the CT image output obtained from both of them. The comparative results for the two CT images, abdomen and thorax are given in Table I. The comparison is illustrated in the Figure 5 which depicts the PSNR of the CT images obtained from the denoising technique with and without the proposed window selection methodology.

TABLE I. PERFORMANCE COMPARISON OF THE DENOISING TECHNIQUE WITH AND WITHOUT THE PROPOSED WINDOW SELECTION METHODOLOGY

| S.No | Noise level (σ) | PSNR values obtained for CT image of Abdomen | | | PSNR values obtained for CT image of Throax | | |
|------|------|------|------|------|------|------|------|
| | | Noisy image | Denoising without proposed methodology | Denoising with proposed methodology | Noisy image PSNR | Denoising without proposed methodology | Denoising with proposed methodology |
| 1 | 10 | 28.13 | 34.67 | 39.81 | 28.15 | 37.31 | 40.24 |
| 2 | 20 | 22.09 | 33.49 | 37.90 | 22.13 | 34.9 | 37.78 |
| 3 | 30 | 18.55 | 33.25 | 36.62 | 18.60 | 33.6 | 35.28 |
| 4 | 40 | 16.09 | 32.76 | 35.39 | 16.10 | 32.9 | 35.15 |
| 5 | 50 | 14.3 | 32.04 | 36.07 | 14.08 | 31.3 | 33.03 |

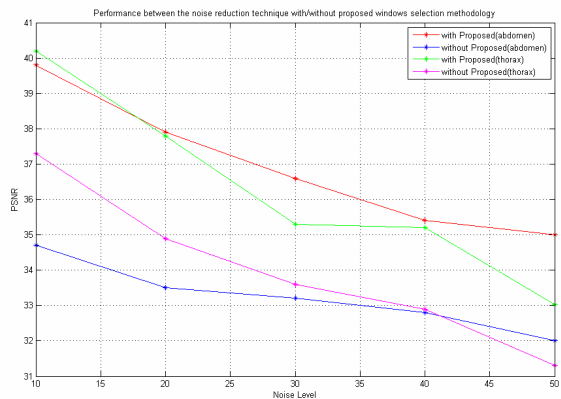

Figure 5. Comparison of the PSNR values obtained from the denoising technique with and without the proposed window selection methodology

In Table I, the PSNR values of the noisy CT images, abdomen and thorax, PSNR values of the denoised image obtained from the denoising technique with and without the proposed methodology are shown. The Figure 5 illustrates the PSNR comparison for the two CT images, Thorax as well as Abdomen. The results show that the PSNR is higher for the denoised image by the denoising technique with the proposed methodology rather than without proposed methodology.

VII. CONCLUSION

In this paper, the proposed GA-based window selection methodology has been described in detail with implementation results. We have proposed the methodology to incorporate the CT image denoising technique using window-based multi-wavelet transformation and thresholding. After the incorporation of the methodology, very closer windows have been obtained for a particular reference window. This has reflected in the performance of the denoising technique. The results have shown that the denoising technique with the proposed window selection methodology provided higher PSNR values rather than the denoising technique without the proposed methodology. This in turns has showed that the denoising technique with the proposed methodology have provided better denoising performance. Hence, by incorporating the proposed methodology in selecting the windows while processing, denoising of CT images can be accomplished in an effective manner as their significance is more.

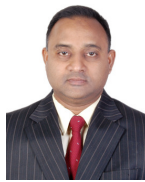

**Prof. Syed Amjad Ali** is a professor and Head in ECE Dept., Lords Institute of Engineering and Technology, Himayathsagar, Hyderabad-8. He is an author of three books: i) Pulse and Digital Circuits ii) Signals and Systems   iii) Digital Signal Processing. He did his M.Tech(Digital Systems and Computer Electronics) from JNTU, Kukatpally,Hyderabad. He has presented the papers in national conferences .He has 16+ years of teaching experience. His area of interest is Image Processing.

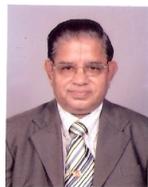

**Dr. Srinivasan Vathsal** was born in Tiruchirapalli, Tamilnadu, India in 1947. He obtained his B.E (Hons), Electrical Engg., in 1968 from Thiagarajar, College of Engineering, Madurai, India, the M.E (Control Systems) in 1970, from BITS, Pilani, India, and the Ph.D in 1974 from I.I.Sc., Bangalore. He worked at the VSSC, Trivandrum, India from 1974-1978 and 1980-1982. During 1982-1984, he was professor of Electrical Engineering at the PSG College of Tech., Coimbatore. He was a senior NRC, NASA Research Associate (1984-1986) at the NASA, GSF Center. He worked at Directorate of SAT of the DRDL in 1988. He was a principal scientist in navigational electronics of Osmania University during 1989-1990, as Head, PFA, directorate of systems in DRDO, Hyderabad during 1990-2005, as Scientist-G, Director, ER & IPR directorate, DRDO, New Delhi. His current research interests are fuzzy logic control, neural networks, missile systems and guidance, radar signal processing and optimal control and image processing. He was awarded a prize for best essay on comprehensive aeronautical policy for India. He is a member of the IEEE, AIAA, Aeronautical Society of India and System Society of India. He has published a number of papers in national and international journals and conferences. Presently working as a Principal in Bhaskar Engineering College, Yenakapally, R.R.Dist., Andhra Pradesh.

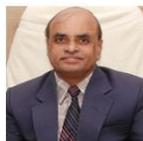

**Dr.K.Lal Kishore** did his B.E. from Osmania University,Hyderabad , M.Tech from Indian Institute of Science, Bangalore,Ph.D from Indian Institute of Science, Bangalore. His Fields of Interest are Micro Electronics and VLSI Engineering.Has more than 25 years of teaching experience. Worked as a Professor and Head in E.C.E department, Chairman, Board of Studies for Electronics and Communication Engineering, Jawaharlal Nehru Technological University, Director, School of Continuing and Distance Education (SCDE) of JNTU Hyderabad, Principal, JNTU College of Engineering, Kukatpally, Hyderabad during June 2002 to 30th June 2004, Director, Academic & Planning (DAP) of JNT University, Director I/c. UGC Academic Staff College of JNT University, Registrar of JNT University.Convener for ECET (FDH) 2000, 2001 and 2005, conducted the Common Entrance Test for Diploma Holders for admission into B.Tech. He has membership in many professional societies like Fellow of IETE, Member IEEE, Member ISTE and Member ISHM. He Won First Bapu Seetharam Memorial Award for Research work from Institution of Electronics and Communication Engineers (IETE), New Delhi in 1986. Received best teacher award from the Government of Andhra Pradesh for the year 2004.He has 31 publications to his credit so far.He wrote three text books i) Electronics Devices and Circuits ii) Electronic Circuit Analysis.   iii) Pulse Circuits.Presently guiding number of Ph.D. students. Now, he is a Rector in JNTU, Kukatpally, Hyderabad.